\documentclass[11pt,a4paper]{article}
\usepackage[hyperref]{emnlp2020}
\usepackage{times}
\usepackage{latexsym}

\usepackage{wasysym}
\usepackage{multirow}
\usepackage{graphicx}

\usepackage{placeins}

\usepackage{microtype}

\aclfinalcopy 
\def\aclpaperid{7} 

\title{Evaluating the Effectiveness of Efficient Neural Architecture Search \\ for Sentence-Pair Tasks}

\author{Ansel MacLaughlin$^{1}$\thanks{Work completed while interning at Amazon.}, Jwala Dhamala$^{2}$, Anoop Kumar$^{2}$, Sriram Venkatapathy$^{2}$,\\ \textbf{Ragav Venkatesan$^{3}$, Rahul Gupta$^{2}$} \\
$^{1}$ Khoury College of Computer Sciences, Northeastern University, Boston, MA \\ $^{2}$ Amazon Alexa, Cambridge, MA \\ $^{3}$ Amazon Alexa, Seattle, WA \\ 
ansel@ccs.neu.edu \quad
  \{jddhamal, anooamzn, vesriram, ragavven, gupra\}@amazon.com
  }

\date{}

\begin{document}
\maketitle

\begin{abstract}

Neural Architecture Search (NAS) methods, which automatically learn entire neural model or individual neural cell architectures, have recently achieved competitive or state-of-the-art (SOTA) performance on variety of natural language processing and computer vision tasks, including language modeling, natural language inference, and image classification. In this work, we explore the applicability of a SOTA NAS algorithm, Efficient Neural Architecture Search (ENAS) \cite{Pham2018EfficientNA} to two sentence pair tasks, paraphrase detection and semantic textual similarity. We use ENAS to perform a micro-level search and learn a task-optimized RNN cell architecture as a drop-in replacement for an LSTM. We explore the effectiveness of ENAS through experiments on three datasets (MRPC, SICK, STS-B), with two different models (ESIM, BiLSTM-Max), and two sets of embeddings (Glove, BERT). In contrast to prior work applying ENAS to NLP tasks, our results are mixed -- we find that ENAS architectures sometimes, but not always, outperform LSTMs and perform similarly to random architecture search.

\end{abstract}

\section{Introduction}
\label{sec:intro}

Neural Architecture Search (NAS) methods aim to automatically discover neural architectures that perform well on a given task and dataset. These methods search over a space of possible model architectures, looking for ones that perform well on the task and will generalize to unseen data. There has been substantial prior work on how to define the architecture search space, search over that space, and estimate model performance \cite{Elsken2019NeuralAS}.  

Recent works, however, cast doubt on the quality and performance of NAS-optimized architectures \cite{Sciuto2020EvaluatingTS,Li2019RandomSA}, showing that current methods fail to find the best performing architectures for a given task and perform similarly to random architecture search.

In this work, we explore applications of a SOTA NAS algorithm, ENAS \cite{Pham2018EfficientNA}, to two sentence-pair tasks, paraphrase detection (PD) and semantic textual similarity (STS). We conduct a large set of experiments testing the effectiveness of ENAS-optimized RNN architectures across multiple models (ESIM, BiLSTM-Max), embeddings (BERT, Glove) and datasets (MRPC, SICK, STS-B). We are the first, to our knowledge, to apply ENAS to PD and STS, to explore applications across multiple embeddings and traditionally LSTM-based NLP models, and to conduct extensive SOTA HPT across multiple ENAS-RNN architecture candidates. 

Our experiments suggest that baseline LSTM models, with appropriate hyperparameter tuning (HPT), can sometimes match or exceed the performance of models with ENAS-RNNs. We also observe that random architectures sampled from the ENAS search space offer a strong baseline, and can sometimes outperform ENAS-RNNs. Given these observations, we recommend that researchers (i) conduct extensive HPT (preferably using automated methods) across various candidate architectures for the fairest comparisons; (ii) compare the performances of ENAS-RNNs against both standard architectures like LSTMs and RNN cells randomly sampled from the ENAS search space; (iii) examine the computational (memory and runtime) requirements of ENAS methods alongside the gains observed.

\section{Related Work}
\label{sec:related}

NAS methods have shown strong performance on many NLP and CV tasks, such as language modeling and image classification \cite{Zoph2017NeuralAS,Pham2018EfficientNA,Luo2018NeuralAO,Liu2019DARTSDA}. Applications in NLP, such as NER \cite{jiang-etal-2019-improved,li-etal-2020-learning}, translation \cite{So2019TheET}, text classification \cite{Wang2020TextNASAN}, and natural language inference (NLI) \cite{pasunuru-bansal-2019-continual-with-url,Wang2020TextNASAN} have also been explored. 

Current SOTA approaches focus on learning new cell architectures as replacements for LSTM or convolutional cells \cite{Zoph2017NeuralAS,Pham2018EfficientNA,Liu2019DARTSDA,jiang-etal-2019-improved,li-etal-2020-learning} or entire model architectures to replace hand-designed models such as the transformer or DenseNet \cite{So2019TheET,Pham2018EfficientNA}.

Recently, the superiority of NAS to random architecture search and traditional architectures with
SOTA HPT methods has been called into question. \citet{Li2019RandomSA} discuss reproducibility issues with current NAS methods and find that, on language modeling and image classification tasks, NAS algorithms perform similarly to random architecture search. Similarly, \citet{Sciuto2020EvaluatingTS} find minimal differences in performance between NAS and random search and that the popular weight-sharing strategy \cite{Pham2018EfficientNA} decreases performance. 
With this in perspective, we conduct a study to investigate the value added by ENAS to two NLP tasks, PD and STS, which, to our knowledge, have not been been explored in previous NAS literature.

\section{Neural Architecture Search for Sentence-Pair Tasks}
\label{sec:nas}

In this work, we explore applications of ENAS to two sentence-pair tasks, PD and STS. We select ENAS because prior work \cite{pasunuru-bansal-2019-continual-with-url,Wang2020TextNASAN} has shown promising results applying it to a closely-related task, NLI, with gains of up to 1.3\% absolute over LSTMs and 1.6\% over an RNN with a random architecture. Through our evaluations on PD and STS, we aim to study whether the ENAS methods used in prior work for NLI are generalizable and whether the results hold when applied to related tasks and datasets.

ENAS models consist of two parts: 1) a search space over model architectures, i.e. child models, and 2) a controller that samples architectures from that search space. The primary contribution of ENAS is that all child models in the search space share their weights, so each child model does not have to be trained from scratch to evaluate it. Training the child models and controller proceeds as follows --  first, the controller is fixed, and the child models are trained together for one epoch on the dataset, sampling a new architecture from the controller to use for each minibatch. Then, the child model shared parameters are fixed, and the controller is updated -- we sample child architectures from its policy and update the controller to maximize the expected reward on the dev set (e.g. dev set accuracy). This two-step process then repeats for a specified number of epochs. After training is complete, a number of child models are sampled from the controller and the best one is trained from scratch and evaluated on the test set. We refer the reader to \citet{Pham2018EfficientNA} for further details on ENAS.

In this work, we follow the setup of \citet{pasunuru-bansal-2019-continual-with-url}, using standard LSTM-based NLP models and replacing the LSTMs with RNN cells sampled from the ENAS controller. We leave the rest of the model architecture (e.g. attention, pooling, output layers) the same, so the child model search space consists of every possible ENAS-RNN architecture with the standard model architecture around it. As with standard ENAS training, the parameters of the ENAS-RNNs and standard model architecture (e.g. final output layer) are shared across all child models.

\subsection{Experiments}
\label{sec:experiments}

We evaluate ENAS on three sentence-pair datasets using two models and two sets of embeddings:

\subsubsection{Sentence-Pair Datasets}
\label{sec:datasets}
\begin{itemize}
    \item \textbf{Microsoft Research Paraphrase Corpus} (MRPC; \citet{Dolan2005AutomaticallyCA}): binary label (sentences are paraphrases or not). 
    \item \textbf{Semantic Textual Similarity Benchmark} (STS-B; \citet{cer-etal-2017-semeval}): similarity score for each sentence-pair in 0 - 5. 
    \item \textbf{SICK-R }\cite{marelli-etal-2014-sick}: similarity score for each sentence-pair in 1 - 5.
\end{itemize} 

\subsubsection{Models}
\label{sec:models}
\begin{itemize}
    \item \textbf{BiLSTM-Max} (BLM, \citet{conneau-etal-2017-supervised}): uses a BiLSTM + max-pooling to form a representation of each sentence ($s_1, s_2$) and forms a joint representation $h = [s_1; s_2; |s_1 - s_2|; s_1 \odot s_2]$. $h$ is then fed through a feedforward layer and a projection to single predicted value. \citet{pasunuru-bansal-2019-continual-with-url} use BLM in their work applying ENAS to NLI.
    \item \textbf{ESIM} \cite{chen-etal-2017-enhanced}: uses 2 BiLSTMs, with a cross-sentence attention module in between, then mean and max pooling get representations of each sentence. It then forms a joint representation $h = [s_{1, avg}; s_{1, max}; s_{2, avg}; s_{2, max}]$ which is fed through a feedforward layer and a final projection to single predicted value.
\end{itemize}

\subsubsection{Embeddings}
\label{sec:embeddings}
\begin{itemize}
    \item \textbf{Feature-based BERT-base} \cite{devlin-etal-2019-bert}: Following \citet{peters-etal-2019-tune}, we jointly encode the sentence pair (rather than encoding each separately). 
    and learn a linear weighted combination of BERT's layers. BERT is frozen during training.  
    \item \textbf{Glove} \cite{pennington-etal-2014-glove}: 300 dimensional vectors trained on Wikipedia and Gigaword. Embeddings are frozen during training.\footnote{Our initial experiments found that static Glove embeddings outperformed non-static ones.}
\end{itemize}

\subsection{LSTM Baselines}
\label{sec:lstm-baselines}
We first benchmark LSTM implementations of both models. We adapt the BLM implementation from \citet{pasunuru-bansal-2019-continual-with-url} and use the AllenNLP implementation of ESIM \cite{Gardner2018AllenNLPAD}. To have the most competitive baselines possible, we perform extensive HPT, running 500 trials using a Tree-structured Parzen Estimator (TPE; \citet{Bergstra2011AlgorithmsFH}). We tune the hidden dimension sizes, dropout rates, batch size, loss function (only for regression tasks: mean squared error or mean absolute error), learning rate, weight decay, grad norm, and random seed. See Appendix~\ref{appendix:hpt} for full HPT experiment details. Note that we put emphasis on extensive, automated HPT and conduct hundreds of HPT trials (as opposed to only tens of trials typically used in prior work, e.g. \citet{yogatama-etal-2015-bayesian}).

Given that we train BLM and ESIM on top of frozen embeddings, we use the ESIM + BERT results from \citet{peters-etal-2019-tune} as a baseline. Our reproduced results are in the same ballpark (Table~\ref{table:dev-test-results}, rows 2-3), albeit with small deviations.

\subsection{ENAS Training} 
\label{sec:enas-training}

After finding the best hyperparameters for each $\langle$dataset, embedding,  model$\rangle$ LSTM configuration, we run ENAS to search for a new RNN for each configuration. Following \citet{pasunuru-bansal-2019-continual-with-url}, we use 6 node ENAS-RNNs. We use Microsoft NNI's \cite{MicrosoftNNI} ENAS implementation. We replace the BiLSTM in BLM and both BiLSTMs in ESIM with the ENAS BiRNNs (we use same architecture in both ESIM layers). We train ENAS for 150 epochs with early-stopping. For each $\langle$dataset, embedding, model) configuration, we train the ENAS models with the same hyperparameters as the best corresponding LSTM model, except learning rate of 1e-4 and grad norm 0.25, which are used across all ENAS models\footnote{Training is unstable with the higher learning rates found during HPT for our LSTM models and those suggested in \citet{pasunuru-bansal-2019-continual-with-url,Pham2018EfficientNA}}. We follow the hyperparameter configurations from \citet{Pham2018EfficientNA} for the ENAS controller. 

\subsection{Training Discovered Architectures}
\label{sec:train-from-scratch}
After training ENAS, we sample 10 architectures from the controller. Just as during ENAS training, we then use these architectures as drop-in replacements for LSTMs, replacing a model's BiLSTM layer(s) with ENAS BiRNN(s). We then train the models from scratch and repeat HPT, extending the original LSTM hyperparameter search space with a choice over the 10 sampled architectures. We run 200 trials of HPT. We note that, unlike the CUDA implementations for LSTMs, it is non-trivial to implement highly optimized arbitrary ENAS-RNN architectures. We discuss these limitations and the overall compute dedicated for HPT on LSTM and ENAS-RNN based models in Appendix~\ref{appendix:hpt}. 

In addition to experiments replacing all BiLSTM layers with ENAS BiRNNs, we also examine mixing ENAS-RNN and LSTM layers in the multi-layer ESIM model. Specifically, we experiment with only replacing the 1st BiLSTM layer in ESIM with an ENAS BiRNN and only replacing the 2nd BiLSTM layer. These models have the same hyperparameter search space as the ESIM model with ENAS-RNNs in both layers (i.e. same possible ENAS-RNN architectures), but we tune and evaluate them separately (see Table~\ref{table:dev-test-results}, rows 5-6, 11-12).

\begin{table*}[ht]
\scriptsize
\centering
\begin{tabular}{rc|ccc|ccc|ccc}
\multicolumn{5}{c}{} & \multicolumn{3}{c}{Dev Performance} & \multicolumn{3}{c}{Test Performance} \\
& Author & Embedding & Model & RNN & SICK-R & MRPC & STS-B & SICK-R & MRPC & STS-B \\ \hline
1. &  \citet{devlin-etal-2019-bert}   & BERT    &    fine-tuned  & -- & -- & -- & -- & \textbf{88.7}      &  \textbf{84.8}    &  \textbf{87.1}     \\
2. & \citet{peters-etal-2019-tune}   & BERT   &  ESIM       & L / L & -- & -- & -- &  86.4     &   78.1   &  82.9   \\ \hline

3. &\multirow{4}{*}{Ours}   & BERT   &  ESIM       & L / L & 88.9 & \textbf{88.0} & 88.0 &  87.0    &   80.2   &  82.0 \\
4. &   & BERT   &  ESIM       & E / E & 88.6 & 87.0 & \textbf{88.5} &  86.8    & 80.8    &  82.2 \\   
5. &   & BERT   &  ESIM       & E / L & \textbf{89.3} & 87.5 & \textbf{88.5} &  \textbf{87.4}    & \textbf{81.0}    &  \textbf{83.0} \\
6. &   & BERT   &  ESIM       & L / E & 88.0 & 87.0 & 88.2 &  86.5    & 79.8    &  81.8   \\ \hline
7. & \multirow{2}{*}{Ours}   & BERT   &  BLM       & L  & 87.4 & 88.0 & 88.1 & 84.8  &  80.4   & 80.1  \\ 
8. &   & BERT   &  BLM       & E &  \textbf{87.8}   &  \textbf{88.5}  & \textbf{88.7} & \textbf{85.5} & \textbf{82.8} & \textbf{83.3}  \\ \hline

9. & \multirow{4}{*}{Ours}   & Glove   &  ESIM       & L / L & 88.6 & \textbf{79.9} & \textbf{83.3} & \textbf{86.1}   & \textbf{73.7} & 75.5  \\
10. &   & Glove   &  ESIM       & E / E & 88.2 & 76.0 & \textbf{83.3} &  85.1   &  69.0   & 75.3    \\
11. &   & Glove   &  ESIM       & E / L & \textbf{88.7} & 77.2 & 83.0 &  85.7   &  71.0   & 72.7    \\
12. &   & Glove   &  ESIM       & L / E & 88.2 & 78.2 & 83.0 &  85.2   &  72.5   & \textbf{76.0}    \\ \hline
13. & \multirow{2}{*}{Ours}   & Glove   &  BLM       & L & 86.3 & \textbf{78.4} & 79.7 & 82.5   &  71.8    & 73.0   \\ 
14. &   & Glove   &  BLM       & E  & \textbf{87.0} & \textbf{78.4} & \textbf{81.6} &  \textbf{84.1}  & \textbf{73.4}   &   \textbf{74.8}  \\ \hline
\end{tabular}
\caption{Dev \& Test set performances for LSTM and ENAS-RNN based models. Following \citet{peters-etal-2019-tune}, we report pearson correlation for SICK-R and STS-B and accuracy for MRPC. In the RNN collumn, ``E'' stands for ENAS-RNN and ``L'' stands for LSTM. For ESIM there can be different of cells in different layers, e.g. E / L stands for ENAS-RNN in the 1st layer and LSTM in the 2nd layer.}
\label{table:dev-test-results}

\begin{tabular}{ccccc|ccc|ccc}
& & & & & \multicolumn{3}{c}{Dev Performance} & \multicolumn{3}{c}{Test Performance} \\

& Embedding & Model & RNN  & RNN Optimized For   & SICK-R  & MRPC & STS-B & SICK-R  & MRPC & STS-B  \\ \hline
1. & BERT & ESIM & E / L & SICK-R          & \textbf{89.3}  & 87.0  & -- & \textbf{87.4} & \textbf{81.3} & -- \\
2. & BERT & ESIM & E / L & MRPC          &  89.0   &  \textbf{87.5} & -- & 86.9 & 81.0 & -- \\
3. & BERT & ESIM &  E / L & STS-B         & --    & --  & \textbf{88.5} & -- & -- & \textbf{83.0} \\
4. & BERT & ESIM & RND / L &    Random     &   88.9  &  87.0   &  88.4  & 87.2 & 79.0 & 81.2  \\ \hline
5. & Glove & BLM & E & SICK-R           & 87.0  & 77.5  & -- & 84.1 & 71.9 & -- \\
6. & Glove & BLM & E &  MRPC          &  87.3   & 78.4  & -- & 83.5 & 73.4 & -- \\
7. & Glove & BLM & E & STS-B         & --    & --  & \textbf{81.6} & -- & -- & \textbf{74.8} \\
8. & Glove & BLM & RND & Random        &  \textbf{87.6}   &  \textbf{79.9}  &  81.1 & \textbf{84.7} & \textbf{75.5} & \textbf{74.8}
\end{tabular}
\caption{Evaluation of how well ENAS-RNNs transfer to other datasets and compare to random search.  We report pearson correlation for SICK-R and STS-B and accuracy for MRPC. In the RNN collumn, ``E'' stands for ENAS-RNN, ``L'' stands for LSTM, and ``RND'' for random RNN. For ESIM we use an ENAS or random RNN in the 1st layer and an LSTM in the 2nd layer.}
\label{table:rnd-comparison}

\end{table*}

\section{Results}
\label{sec:results}

Table~\ref{table:dev-test-results} lists the dev and test results for all datasets, embeddings, and models. We focus our discussion on the test results. On the whole, the results are mixed. $\langle$BLM, ENAS$\rangle$ outperforms $\langle$BLM, LSTM$\rangle$ across all datasets and embeddings by an average of 1.9\%. $\langle$ESIM, ENAS$\rangle$, on the other hand, fails to consistently outperform $\langle$ESIM, LSTM$\rangle$. ESIM models with ENAS-RNNs in both layers lag behind LSTMs by 0.9\%, on average. 

Focusing first on BLM, we find that $\langle$BLM, ENAS$\rangle$ outperforms $\langle$BLM, LSTM$\rangle$ by an average of 2.1\% across all three datasets using BERT (row 8) and 1.7\% using Glove (row 14). These results parallel those of \citet{pasunuru-bansal-2019-continual-with-url}, who find that $\langle$BLM, ENAS$\rangle$ with ELMO embeddings \cite{peters-etal-2018-deep} outperforms $\langle$BLM, LSTM$\rangle$ on two NLI datasets and is on par on a third. However, both in our experiments and those of \citet{pasunuru-bansal-2019-continual-with-url}, the 6 node ENAS-RNNs have more parameters than the corresponding LSTM models\footnote{The exact ratio in number of parameters between 6 node ENAS-RNNs and LSTMs depends on the input and hidden dimensions}, making it difficult to get a clear picture of the effects of just changing the RNN architecture. To control for this, in \S\ref{sec:random-transfer} we conduct experiments comparing ENAS-RNNs to RNNs randomly sampled from the same search space.

Examining ESIM, the results are mixed. ESIM models with ENAS-RNNs in both layers (rows 4, 10) are worse than $\langle$ESIM, LSTM$\rangle$ on 4 of 6 $\langle$dataset, embedding$\rangle$ configurations. The best $\langle$ESIM, ENAS$\rangle$ performance is actually achieved using a mix of ENAS-RNNs and LSTMs across different layers. In fact, the only configurations in which $\langle$ESIM, ENAS$\rangle$ outperforms $\langle$ESIM, LSTM$\rangle$ across all three datasets is $\langle$BERT, ENAS / LSTM) (row 5), where we only replace the first LSTM layer with an ENAS-RNN. The gains, however, are modest compared to those of the BLM model, improving over $\langle$ESIM, LSTM$\rangle$ by 0.73\% on average. Further, changing the embeddings to Glove $\langle$Glove, ENAS / LSTM) (row 11), $\langle$ESIM, ENAS$\rangle$ underperforms $\langle$ESIM, LSTM$\rangle$ across all 3 datasets by nearly 2\% on average. Since we do not observe similar performance gains with ESIM as with BLM, we hypothesize that optimization of specific RNN architectures might matter less as model complexity (e.g. number of layers) increases. We suggest future work further examine the importance of ENAS as it relates to model complexity, especially on tasks where an RNN's architecture might have a higher impact on modeling performance.

\subsection{Random \& Transfer Architectures}
\label{sec:random-transfer}

In addition to comparisons to LSTMs, we evaluate two common claims about NAS methods: 1) NAS outperforms random search \cite{Pham2018EfficientNA,Zoph2017NeuralAS,Luo2018NeuralAO,Liu2019DARTSDA} 2) NAS architectures are transferable to related datasets and tasks \cite{Zoph2017NeuralAS,Liu2019DARTSDA,Luo2018NeuralAO}. We choose two configurations to evaluate these claims: (i) $\langle$Glove, BLM$\rangle$ and (ii) $\langle$BERT, ESIM, ENAS / LSTM$\rangle$ with ENAS-RNNs only in the first layer, keeping the second BiLSTM layer. We chose these configurations since they perform well relative to LSTMs and, between them, cover all embeddings and models.

For claim \#1, we first randomly sample 10 RNN architectures from the ENAS search space. Then, just as for the ENAS-RNNs, we perform 200 HPT trials, replacing the 10 ENAS-RNN candidates with the 10 randomly sampled RNN candidates. For claim \#2, we test the transferability of SICK-R and MRPC cells to/from each other. We do not evaluate the transferability of STS-B cells, since STS-B contains data from SICK-R and MRPC. We again perform 200 HPT trials, but with the different dataset's ENAS-RNN cells in the search space.

Table~\ref{table:rnd-comparison} shows our results. We again focus on test results. For claim \#1, we find mixed results, with ENAS outperforming random search by an average of 1.33\% in the configuration $\langle$BERT, ESIM, ENAS / LSTM$\rangle$ (rows 1-4), but performing worse or on par with random on $\langle$GLOVE, BLM$\rangle$ (rows 5-8) (average 0.9\% decrease). These results contrast those of \citet{Pham2018EfficientNA,pasunuru-bansal-2019-continual-with-url}, who report gains over random search on language modeling (25.4\% decrease in perplexity) and NLI datasets (1.53\% increase in accuracy). We hypothesize that these differences are due, in part, to our emphasis on creating strong baselines by searching over multiple architectures and performing extensive HPT for all models and settings. 

For claim \#2, we find that transfer architectures underperform dataset-specific ENAS architectures by 0.58\% and random architectures by 0.7\%, on average. Only one architecture (row 1, SICK to MRPC) outperforms either of the corresponding random or dataset-specific architectures. Together with our findings for claim \#1, these results cast further doubt on the ability of ENAS to find the best architecture for a specific task, its superiority to well-tuned random architectures, and the transferability of its discovered architectures.

\section{Conclusion}
\label{sec:conclusion}

Unlike prior work applying ENAS to NLP, we find that ENAS-RNNs only outperform LSTMs and random search on some $\langle$dataset, embedding, model) configurations. Our findings parallel recent work \cite{Li2019RandomSA,Sciuto2020EvaluatingTS} which question the effectiveness of current NAS methods and their superiority to random architecture search and SOTA HPT methods. 
Given our mixed results, we recommend researchers: (i) extensively tune hyperparameters for standard (e.g. LSTM) and randomly sampled architectures to create strong baselines; (ii) benchmark ENAS performance across multiple simple and complex model architectures (e.g. BLM \& ESIM); (iii) present computational requirements alongside gains observed with ENAS methods.

\bibliographystyle{acl_natbib}
\bibliography{anthology,emnlp2020}

\appendix

\section{Implementation Details}

All models were implemented with Pytorch and run on Amazon p3 instances (16GB Nvidia V100).

\subsection{Embeddings}
 Experiments with BERT used the Huggingface Transformers library \citep{Wolf2019HuggingFacesTS}. Experiments with Glove vectors used 300 dimensional vectors trained on Wikipedia 2014 + Gigaword 5\footnote{\url{http://nlp.stanford.edu/data/glove.6B.zip}}. Glove vectors weren't updated training, and out-of-vocabulary tokens were replaced with the token ``[UNK]'' with an embedding of all 0s ($\approx 6\%$ of tokens are OOV). In initial experiments, we found no differences between our all-0 embeddings and embeddings randomly initialized according to a Gaussian distribution.   

\subsection{Hyperparameter Tuning}
\label{appendix:hpt}
All HPT was run using Microsoft NNI's parallel implementation of TPE\footnote{\url{https://nni.readthedocs.io/en/latest/CommunitySharings/ParallelizingTpeSearch.html}} with concurrency 8. Table~\ref{table:hpt-search-space} contains the search space for our experiments. Table~\ref{table:hpt-values} contains the best hyperparameter settings for all of our experiments.

\begin{table}
\scriptsize
\centering
\begin{tabular}{l|l}
\textbf{Hyperparameter} & \textbf{Search Space} \\ \hline
 batch size     &  [16, 32, 64]            \\ \hline
  learning rate             &  0.0001 - 0.01 \\\hline
  loss function & classification: cross entropy \\  
                & regression: [mae, mse] \\ \hline
  weight decay & 0.001 - 0.1 \\  \hline
  grad norm & 0.25 - 20.0 \\  \hline
  hidden dimensions & w/ bert: [384, 512, 768, 1152, 1536] \\  
                    & w/ glove: [150, 200, 300, 450, 600]  \\ \hline
  dropouts & 0.25 - 0.75 \\ \hline
  random seed & [0, 1, 2, 3, 4, 5] \\ \hline
  epochs & 75 (with early stopping) \\ \hline
  RNN Architecture & Choice of 10 unique architectures \\ & (only for models with ENAS-RNNs) 
\end{tabular}
\caption{Hyperparameter search space for all experiments.}
\label{table:hpt-search-space}
\end{table}

\subsubsection{Memory Limitations for HPT with ENAS-RNNs}
In order for a model to fit on a single GPU (16GB Nvidia V100), we had to limit the search space slightly for models using both ENAS-RNNs and BERT embeddings. This is because the ENAS-RNN search space contains weight matrices $W^h_{\ell, j}$ between each pair of nodes $\ell, j$ in the RNN search space DAG, which greatly expands memory usage (see \citet{Pham2018EfficientNA}, 
sections 2.1 and Appendix A). For both BLM and ESIM models, hidden dimensions were limited to [384, 512, 768]. Further, for ESIM models with ENAS-RNNs in both layers, the batch size was also limited to [16, 32].

\subsubsection{Timing limitations for HPT with ENAS-RNNs}
Since our ENAS-RNNs are, similar to prior NAS research code, implemented using a Python for-loop over time steps, the implementation is significantly slower ($\approx$ 25x) than the cuda-optimized LSTM equivalent. Thus, due to computational limits, we only perform 200 trials of HPT for the models with ENAS-RNNs (vs. 500 for models with LSTMs). Though the number of HPT trials is lower than for LSTMs, due to their slow speed, the total compute time devoted to tuning the ENAS-RNN models is roughly 10x+ higher. As an example, Table~\ref{table:hpt-cost} shows the total compute time dedicated to HPT for BLM models (both LSTM-based models and ENAS-RNN based models), measured as the total number of hours spent on a single p3.16xlarge instance\footnote{\url{https://aws.amazon.com/ec2/instance-types/p3/}} to finish all HPT trials. Note, the models with ENAS-RNNs are not always exactly 10x slower than the LSTM equivalent -- since we are also searching over batch size during HPT, runtimes can vary significantly.

\begin{table}
\scriptsize
\centering
\begin{tabular}{c|c|c}
Configuration & \# HPT Trials & Runtime (hours) \\ \hline
(BLM, BERT, LSTM, SICK) & 500 & 19.14  \\ 
(BLM, BERT, LSTM, MRPC) & 500 & 23.99 \\ 
(BLM, BERT, LSTM, STS-B) & 500 & 47.30 \\ \hline
(BLM, BERT, ENAS, SICK) & 200 & 79.21 \\ 
(BLM, BERT, ENAS, MRPC) & 200 & 67.29 \\ 
(BLM, BERT, ENAS, STS-B) & 200 & 141.09 \\ \hline 

(BLM, Glove, LSTM, SICK) & 500 & 5.35 \\ 
(BLM, Glove, LSTM, MRPC) & 500 & 4.90 \\ 
(BLM, Glove, LSTM, STS-B) & 500 & 9.36 \\ \hline
(BLM, Glove, ENAS, SICK) & 200 & 78.65 \\ 
(BLM, Glove, ENAS, MRPC) & 200 & 113.33 \\ 
(BLM, Glove, ENAS, STS-B) & 200 & 193.93 \\ \hline 

\end{tabular}
\caption{Compute time spent on HPT for BLM models (both LSTM-based models and ENAS-RNN based models). Compute time measured as total number of hours on a single p3.16xlarge instance. All HPT was run using Microsoft NNI's parallel implementation of TPE\footnote{\url{https://nni.readthedocs.io/en/latest/CommunitySharings/ParallelizingTpeSearch.html}} with concurrency 8 (one trial running on each of the 8 GPUs in the p3.16xlarge instance).}
\label{table:hpt-cost}
\end{table}

\subsection{Memory Limitations for Training ENAS}
As noted in \S\ref{sec:enas-training},
we train the ENAS child models $\langle$BLM, ESIM$\rangle$ using the same parameters as the corresponding best LSTM model for the given configuration $\langle$dataset, embeddings, model$\rangle$. For the configuration $\langle$STS-B, BERT, ESIM$\rangle$, the corresponding ENAS child models would not fit on a single GPU (16GB Nvidia V100). This is due to the large memory footprint of ENAS as discussed in \ref{appendix:hpt}. Thus, for $\langle$STS-B, BERT, ESIM$\rangle$ we decrease the batch size from 64 to 32 and the hidden dimensions from 1152 to 768. 

\subsection{ESIM: Differences Between Training Child Models with ENAS and Training Models from Scratch}
\label{appendix:esim_differences}
As described in \S\ref{sec:enas-training}, when training the ESIM child models jointly with the ENAS controller, we replace both of ESIM's BiLSTMs with the sampled ENAS-RNN architectures. We do this for each $\langle$dataset, embedding$\rangle$ configuration, thus running 6 total instances of ENAS (3 datasets * 2 embeddings). After the ENAS training is complete, we sample 10 ENAS-RNN architectures from the trained controller. 

However, when training ESIM models from scratch, as described in \S\ref{sec:train-from-scratch}, 
we experiment with 1) replacing both LSTM layers with the ENAS-RNN architecture (same as during ENAS training) 2) only replacing the 1st layer 3) only replacing the 2nd layer. We treat each ESIM layer configuration as its own model and tune its hyperparameters separately. Thus, for example, for the configuration (SICK-R, BERT, ESIM) we perform 200 trails of HPT for the configuration with ENAS-RNNs in both layers, 200 trials of HPT for the configuration with an ENAS-RNN in layer 1 and an LSTM in layer 2, and finally 200 trials of HPT for the configuration with an LSTM in layer 1 and an ENAS-RNN in layer 2. Note, however, that these three separate instances of HPT share the same search space over ENAS-RNN architectures -- all three are searching over the same 10 ENAS-RNNs sampled from the same controller. In total, we run 18 different instances of HPT (3 datasets * 2 embeddings * 3 layer configs). The results from each configuration are presented separately in Table~1 (in the main portion of the paper).

\subsection{RNN Architectures Sampled from ENAS Search Space}
Table~\ref{table:rnn_archs} shows the architectures of all RNNs used in our experiments (ENAS-RNNs, transferred ENAS-RNNs, random RNNs). Each architecture is numbered 1-26. Table~\ref{table:hpt-values}, which displays the hyperparameter settings for each model and configuration, lists which RNN architecture each configuration uses. 

Note, some of the architectures are the same across different model configurations. This is due to two reasons: 

\begin{itemize}
    \item As discussed in \S\ref{sec:train-from-scratch} and \S\ref{appendix:esim_differences}, we experiment with mixing ENAS-RNN and LSTM layers in the multi-layer ESIM model. The ESIM models with ENAS RNNs in both layers share the same possible ENAS-RNN architectures as the corresponding ESIM models with an ENAS-RNN only in the 1st layer or 2nd layer. 
    \item We sampled 10 total random architectures from the ENAS-RNN search space then used those same 10 architectures in the search spaces for all $\langle$dataset, model, embedding$\rangle$ configurations. Thus, some configurations might use the same architecture.
\end{itemize}

\subsection{Datasets}
For MRPC and STS-B, we use the data provided by Glue\footnote{\url{https://gluebenchmark.com/faq}}. For SICK-R, we use the data provided by SemEval-2014 Task 1\footnote{\url{http://alt.qcri.org/semeval2014/task1/}}. We use scikit-learn\footnote{\url{https://scikit-learn.org/stable/modules/generated/sklearn.model_selection.train_test_split.html}, dev size: 0.1, random state: 0} to split the provided SICK-R training data into train and dev splits.

For our experiments with BERT, we use the BertTokenizer from the Huggingface Transformers library \citep{Wolf2019HuggingFacesTS}. We cap each sentence-pair at a certain number of total wordpiece tokens (SICK: 64, MRPC: 128, STS-B: 128). For our experiments with Glove, we use spacy\footnote{\url{https://spacy.io/models/en\#en\_core\_web\_md}} \citep{spacy2} to tokenize each sentence. We cap each sentence at a certain number of tokens (SICK: 30, MRPC: 46, STS-B: 39).

\begin{table*}
\tiny
\centering
\setlength\tabcolsep{1pt}
\begin{tabular}{cccc|cccccccccc}
Model & Embedding & RNN & Dataset & Batch Size & Learning Rate & Loss & Weight Decay & Grad Norm & Hidden Dim & Dropout & Variational Dropout & Rnd Seed & Architecture \# \\ \hline

BLM & BERT & L & SICK & 32 & 0.0046 & mse & 0.0514 & 12.3656 & 512 & (0.3782, 0.3474) & 0.4088 & 3 & --
 \\ 
BLM & BERT & L & MRPC & 64 & 0.0021 & cross entropy & 0.0637 & 12.8279 & 384 & (0.6355, 0.4388) & 0.6804 & 2 & --

 \\ 
BLM & BERT & L & STS-B & 32 & 0.0075 & mse & 0.0407 & 16.8742 & 512 & (0.2702, 0.4525) & 0.6783 & 2 & --
 \\ \hline

BLM & Glove & L & SICK & 64 & 0.0007 & mse & 0.0040 & 10.2636 & 300 & (0.3555, 0.2937) & 0.2774 & 1 & --
 \\ 
BLM & Glove & L & MRPC & 32 & 0.0017 & cross entropy & 0.0301 & 8.1649 & 450 & (0.3346, 0.3751) & 0.2986 & 5 & --
 \\ 
BLM & Glove & L & STS-B & 32 & 0.0004 & mse & 0.0201 & 4.9461 & 200 & (0.2597, 0.5924) & 0.4516 & 0 & --
 \\ \hline

BLM & BERT & E & SICK & 32 & 0.0074 & mse & 0.0226 & 11.2817 & 384 & (0.3372, 0.5304) & 0.3009 & 5 & 17
\\ 
BLM & BERT & E & MRPC & 32 & 0.0031 & cross entropy & 0.0670 & 9.4340 & 384 & (0.5310, 0.6235) & 0.4676 & 1 & 15\\ 
BLM & BERT & E & STS-B & 32 & 0.0019 & mae & 0.0382 & 6.7670 & 512 & (0.2507, 0.4492) & 0.6193 & 1 & 19
 \\ \hline

BLM & Glove & E & SICK & 64 & 0.0007 & mse & 0.0729 & 11.7080 & 450 & (0.3199, 0.2711) & 0.3911 & 5 & 18
 \\ 
BLM & Glove & E & MRPC & 64 & 0.0001 & cross entropy & 0.0637 & 15.5210 & 450 & (0.3352, 0.3993) & 0.2948 & 4 & 16
\\ 
BLM & Glove & E & STS-B & 16 & 0.0007 & mae & 0.0258 & 2.9847 & 450 & (0.2584, 0.6419) & 0.2508 & 4 & 20
 \\ \hline

BLM & Glove & R & SICK & 64 & 0.0016 & mse & 0.0647 & 15.0969 & 450 & (0.2505, 0.3945) & 0.2589 & 0 & 24
 \\ 
BLM & Glove & R & MRPC & 64 & 0.0015 & cross entropy & 0.0956 & 12.2487 & 300 & (0.2956, 0.3971) & 0.3304 & 0 & 22
 \\ 
BLM & Glove & R & STS-B & 64 & 0.0004 & mse & 0.0257 & 1.2826 & 600 & (0.3355, 0.4312) & 0.3392 & 3 & 24
 \\ \hline

BLM & Glove & T & SICK & 32 & 0.0003 & mse & 0.0058 & 6.1308 & 300 & (0.3809, 0.3487) & 0.3273 & 2 & 25\\ 
BLM & Glove & T & MRPC & 32 & 0.0005 & cross entropy & 0.0341 & 14.0270 & 200 & (0.4586, 0.6012) & 0.4123 & 0 & 26\\ \hline

ESIM & BERT & L/L & SICK & 32 & 0.0011 & mae & 0.0299 & 12.9599 & (512 1152) & (0.3171, 0.6050) & (0.6962, 0.4123) & 4 & --
\\ 
ESIM & BERT & L/L & MRPC & 64 & 0.0048 & cross entropy & 0.0448 & 16.0686 & (384 512) & (0.2806, 0.4960) & (0.5453, 0.3357) & 1 & --
 \\ 
ESIM & BERT & L/L & STS-B & 64 & 0.0011 & mae & 0.0855 & 18.4787 & (1152 1152) & (0.4213, 0.4769) & (0.5011, 0.5806) & 3 & --
 \\ \hline

ESIM & Glove & L/L & SICK & 32 & 0.0018 & mse & 0.0804 & 12.3511 & (200 300) & (0.4369, 0.5705) & (0.4491, 0.3239) & 1 & --
 \\ 
ESIM & Glove & L/L & MRPC & 64 & 0.0006 & cross entropy & 0.0415 & 16.6595 & (600 200) & (0.4089, 0.7434) & (0.2795, 0.4438) & 3 & --
\\ 
ESIM & Glove & L/L & STS-B & 64 & 0.0027 & mse & 0.0741 & 12.3487 & (300 600) & (0.2822, 0.4862) & (0.2867, 0.5283) & 1 & --
 \\ \hline

ESIM & BERT & E/E & SICK & 16 & 0.0002 & mae & 0.0572 & 4.5861 & (512 768) & (0.3362, 0.6338) & (0.6415, 0.3806) & 1 & 7
 \\ 
ESIM & BERT & E/E & MRPC & 32 & 0.0005 & cross entropy & 0.0808 & 15.6688 & (384 768) & (0.7098, 0.6014) & (0.6504, 0.3573) & 2 & 5
 \\ 
ESIM & BERT & E/E & STS-B & 32 & 0.0024 & mse & 0.0684 & 17.1467 & (384 512) & (0.4992, 0.6578) & (0.7135, 0.4686) & 5 & 12
 \\ \hline

ESIM & Glove & E/E & SICK & 64 & 0.0005 & mse & 0.0673 & 11.2588 & (450 200) & (0.5421, 0.6383) & (0.4262, 0.4960) & 1 & 11
 \\ 
ESIM & Glove & E/E & MRPC & 64 & 0.0019 & cross entropy & 0.0544 & 16.2351 & (150 600) & (0.4805, 0.6752) & (0.4711, 0.5483) & 3 & 6
 \\ 
ESIM & Glove & E/E & STS-B & 64 & 0.0005 & mae & 0.0579 & 11.3040 & (450 200) & (0.3348, 0.5270) & (0.2846, 0.4997) & 0 & 13
 \\ \hline

ESIM & BERT & E/L & SICK & 16 & 0.0008 & mse & 0.0835 & 14.1718 & (512 768) & (0.3996, 0.4231) & (0.3149, 0.3665) & 0 & 9
 \\ 
ESIM & BERT & E/L & MRPC & 32 & 0.0005 & cross entropy & 0.0525 & 13.0402 & (768 512) & (0.5491, 0.2819) & (0.4482, 0.3430) & 5 & 3
 \\ 
ESIM & BERT & E/L & STS-B & 32 & 0.0008 & mae & 0.0995 & 5.6442 & (384 384) & (0.6291, 0.6221) & (0.3899, 0.6917) & 5 & 14
 \\ \hline

ESIM & Glove & E/L & SICK & 32 & 0.0004 & mse & 0.0337 & 0.7994 & (600 600) & (0.4193, 0.6904) & (0.4331, 0.6221) & 2 & 10
 \\ 
ESIM & Glove & E/L & MRPC & 64 & 0.0011 & cross entropy & 0.0549 & 5.7392 & (200 150) & (0.5909, 0.4142) & (0.4288, 0.2503) & 4 & 4
 \\ 
ESIM & Glove & E/L & STS-B & 64 & 0.0003 & mse & 0.0302 & 13.5390 & (450 600) & (0.4538, 0.2828) & (0.4641, 0.6847) & 0 & 13
 \\ \hline

ESIM & BERT & R/L & SICK & 64 & 0.0007 & mse & 0.0135 & 3.3407 & (384 512) & (0.3738, 0.4779) & (0.6879, 0.3507) & 2 & 23
 \\ 
ESIM & BERT & R/L & MRPC & 64 & 0.0007 & cross entropy & 0.0747 & 12.8833 & (384 768) & (0.3532, 0.6506) & (0.6440, 0.6599) & 0 & 21
\\ 
ESIM & BERT & R/L & STS-B & 32 & 0.0014 & mse & 0.0240 & 0.3344 & (512 384) & (0.6102, 0.2993) & (0.5616, 0.3264) & 4 & 24
 \\ \hline

ESIM & BERT & T/L & SICK & 64 & 0.0025 & mse & 0.0623 & 6.0643 & (384 384) & (0.4455, 0.3305) & (0.6036, 0.4636) & 3 & 5
 \\ 
ESIM & BERT & T/L & MRPC & 32 & 0.0003 & cross entropy & 0.0989 & 19.2888 & (512 768) & (0.3023, 0.2515) & (0.6723, 0.4313) & 3 & 7
 \\ \hline

ESIM & BERT & L/E & SICK & 32 & 0.0024 & mse & 0.0690 & 6.5209 & (384 384) & (0.2935, 0.3905) & (0.5975, 0.3623) & 2 & 7
 \\ 
ESIM & BERT & L/E & MRPC & 32 & 0.0020 & cross entropy & 0.0637 & 12.9123 & (768 768) & (0.3302, 0.5489) & (0.7050, 0.5593) & 0 & 1
 \\ 
ESIM & BERT & L/E & STS-B & 16 & 0.0014 & mae & 0.0294 & 19.7594 & (384 384) & (0.3857, 0.5279) & (0.5551, 0.3715) & 3 & 12
 \\ \hline

ESIM & Glove & L/E & SICK & 32 & 0.0028 & mse & 0.0360 & 16.7776 & (150 200) & (0.3367, 0.7101) & (0.3469, 0.3811) & 3 & 8
\\ 
ESIM & Glove & L/E & MRPC & 64 & 0.0013 & cross entropy & 0.0151 & 3.7091 & (300 300) & (0.4849, 0.6060) & (0.5526, 0.4104) & 0 & 2
 \\ 
ESIM & Glove & L/E & STS-B & 32 & 0.0017 & mse & 0.0814 & 0.2999 & (150 200) & (0.2829, 0.3279) & (0.2622, 0.2951) & 5 & 13
 \\ \hline

\end{tabular}
\caption{Hyperparameter values used for all experiments. In the RNN collumn, ``E'' stands for ENAS-RNN, ``L'' stands for LSTM, ``R'' for random RNN, and ``T'' for transfer. All floating point values have been rounded to 4 significant figures after the decimal point. Variational dropout is applied before each RNN layer. For models with RNNs from the ENAS search space (all models except those with LSTMs), the column `Architecture \#' displays which RNN architecture it uses. The number corresponds to the row number in Table~\ref{table:rnn_archs}. For ESIM models, the two hidden dimension values refer to (RNN layer 1, RNN layer 2) and the two dropout numbers refer to standard dropout (applied after the 'enhancement' layer, in the final MLP layer). For BLM models, the two dropout numbers refer to standard dropout applied (after the RNN layer, before the final projection)}
\label{table:hpt-values}
\end{table*}

\begin{table*}
\tiny
\centering
\setlength\tabcolsep{3pt}
\begin{tabular}{c|ccccccccccc}
& Node 0 Op &  Node 1 Input &  Node 1 Op & Node 2 Input &  Node 2 Op & Node 3 Input &  Node 3 Op & Node 4 Input &  Node 4 Op & Node 5 Input &  Node 5 Op  \\ \hline

1. & Tanh & 0 & Relu & 0 & Relu & 0 & Relu & 0 & Relu & 0 & Relu \\ 
2. & Tanh & 0 & Relu & 1 & Relu & 2 & Relu & 0 & Relu & 2 & Relu \\ 
3. & Tanh & 0 & Relu & 1 & Relu & 0 & Identity & 0 & Identity & 0 & Identity \\
4. & Identity & 0 & Relu & 0 & Sigmoid & 0 & Relu & 2 & Relu & 1 & Relu \\
5. & Tanh & 0 & Relu & 0 & Relu & 0 & Identity & 0 & Identity & 4 & Relu \\
6. & Identity & 0 & Sigmoid & 0 & Relu & 0 & Relu & 2 & Relu & 3 & Relu \\
7. & Tanh & 0 & Tanh & 0 & Relu & 0 & Tanh & 3 & Tanh & 0 & Tanh \\
8. & Tanh & 0 & Identity & 0 & Tanh & 0 & Identity & 0 & Identity & 0 & Tanh \\
9. & Tanh & 0 & Tanh & 0 & Relu & 0 & Tanh & 0 & Tanh & 0 & Tanh \\
10. & Tanh & 0 & Identity & 0 & Tanh & 0 & Identity & 0 & Tanh & 0 & Identity \\
11. & Tanh & 0 & Tanh & 0 & Identity & 0 & Tanh & 0 & Identity & 0 & Identity \\
12. & Relu & 0 & Tanh & 1 & Sigmoid & 0 & Relu & 0 & Sigmoid & 0 & Relu \\
13. & Identity & 0 & Identity & 1 & Identity & 0 & Sigmoid & 3 & Identity & 0 & Sigmoid \\
14. & Sigmoid & 0 & Relu & 0 & Sigmoid & 0 & Relu & 0 & Relu & 0 & Sigmoid \\
15. & Sigmoid & 0 & Identity & 0 & Tanh & 0 & Tanh & 0 & Tanh & 0 & Tanh \\
16. & Sigmoid & 0 & Tanh & 0 & Tanh & 0 & Identity & 0 & Identity & 0 & Identity \\
17. & Tanh & 0 & Sigmoid & 1 & Sigmoid & 2 & Relu & 3 & Sigmoid & 1 & Sigmoid \\
18. & Tanh & 0 & Sigmoid & 1 & Sigmoid & 0 & Sigmoid & 1 & Sigmoid & 2 & Sigmoid \\
19. & Sigmoid & 0 & Sigmoid & 0 & Sigmoid & 0 & Relu & 0 & Relu & 0 & Sigmoid \\
20. & Relu & 0 & Sigmoid & 1 & Sigmoid & 0 & Sigmoid & 0 & Sigmoid & 0 & Sigmoid \\
21. & Tanh & 0 & Identity & 0 & Sigmoid & 1 & Tanh & 2 & Sigmoid & 0 & Tanh \\
22. & Sigmoid & 0 & Relu & 0 & Sigmoid & 0 & Sigmoid & 2 & Identity & 0 & Identity \\
23. & Tanh & 0 & Sigmoid & 0 & Relu & 0 & Relu & 2 & Tanh & 1 & Identity \\
24. & Sigmoid & 0 & Sigmoid & 1 & Tanh & 1 & Sigmoid & 1 & Sigmoid & 1 & Relu \\
25. & Sigmoid & 0 & Tanh & 0 & Tanh & 0 & Identity & 0 & Tanh & 1 & Identity \\
26. & Identity & 0 & Sigmoid & 0 & Identity & 0 & Sigmoid & 0 & Sigmoid & 4 & Sigmoid \\

\end{tabular}
\caption{RNN Architectures (from the ENAS RNN search space) used across all experiments (including ENAS-RNNs, random RNNs and transfer architectures). These architectures are matched with their corresponding model configuration in Table~\ref{table:hpt-values} by the column `Architecture \#'. Node\_\#\_Input refers to the index of the previous node used as input to the current node. Node\_\#\_Op refers to the elementwise operation applied at each node (Relu, Tanh, Sigmoid, Identity). Please see \citet{Pham2018EfficientNA} for more details on the ENAS RNN search space.}
\label{table:rnn_archs}
\end{table*}

\end{document}